\documentclass{mlr_template/applemlr}

\usepackage{amsmath}
\usepackage{enumerate}
\usepackage{algorithm}
\usepackage{algpseudocode}
\usepackage{amsfonts}
\usepackage{amsthm}
\usepackage{cleveref}
\usepackage{diagbox}
\usepackage{colortbl}
\usepackage{amssymb}
\usepackage{xspace}
\usepackage{wrapfig}
\usepackage{adjustbox}
\usepackage{tabularx}
\usepackage{booktabs}
\usepackage{mathtools}
\usepackage{tikz}
\usepackage{enumitem}
\usepackage{silence}
\usepackage{dsfont}
\usepackage{multirow}
\usepackage{makecell}
\usepackage{xfakebold}
\usepackage{placeins} %
\usepackage{float}
\usepackage{flafter}

\definecolor{textgray}{HTML}{6E6E73}
\usetikzlibrary{positioning, calc}
\usetikzlibrary{decorations.pathmorphing}

\makeatletter
\patchcmd{\wrong@fontshape}{\@gobbletwo}{}{}{}
\makeatother
\numberwithin{equation}{section}
\makeatletter
\AtBeginDocument{
  \urlstyle{sf}
  
}
\makeatother

\definecolor{light}{RGB}{125, 125, 125}
\crefname{tcb@cnt@pbox}{code}{code}
\Crefname{tcb@cnt@pbox}{Code}{Code}
\crefname{assumption}{assumption}{assumption}
\Crefname{assumption}{Assumption}{Assumptions}

\newtcolorbox[auto counter]{pbox}[2][]{
  colback=white,
  title=Code~\thetcbcounter: #2,
  #1,fonttitle=\sffamily,
  fontupper=\sffamily,
  arc=2pt,
  colframe=bgcolor,
  coltitle=fgcolor,
  colbacktitle=bgcolor,
  toptitle=0.25cm,
  bottomtitle=0.125cm
}

\makeatletter
\newcommand\applefootnote[1]{%
  \begingroup
  \renewcommand\thefootnote{}%
  \renewcommand\@makefntext[1]{\noindent##1}%
  \footnote{#1}%
  \addtocounter{footnote}{-1}%
  \endgroup
}
\makeatother

\definecolor{cverbbg}{gray}{0.90}

\usepackage{fontspec}
\usepackage{xeCJK}
\usepackage{fontspec}
\usepackage{polyglossia}
\usepackage{microtype}
\usepackage{graphicx}
\usepackage{booktabs,longtable,array,colortbl}
\usepackage{etoolbox}

\usepackage{xeCJK}
\usepackage{polyglossia}
\setmainlanguage{english}
\setotherlanguages{french,russian,japanese,chinese}

\makeatother

\makeatletter
\patchcmd{\hyper@makecurrent}{%
    \ifx\Hy@param\Hy@chapterstring
        \let\Hy@param\Hy@chapapp
    \fi
}{%
    \iftoggle{inappendix}{%
        \@checkappendixparam{chapter}%
        \@checkappendixparam{section}%
        \@checkappendixparam{subsection}%
        \@checkappendixparam{subsubsection}%
        \@checkappendixparam{paragraph}%
        \@checkappendixparam{subparagraph}%
    }{}%
}{}{\errmessage{failed to patch}}

\newcommand*{\@checkappendixparam}[1]{%
    \def\@checkappendixparamtmp{#1}%
    \ifx\Hy@param\@checkappendixparamtmp
        \let\Hy@param\Hy@appendixstring
    \fi
}
\makeatletter

\newtoggle{inappendix}
\togglefalse{inappendix}

\apptocmd{\appendix}{\toggletrue{inappendix}}{}{\errmessage{failed to patch}}

\newcommand{\numlangs}{14}

\title{Multilingual Reasoning Gym: Multilingual Scaling of Procedural Reasoning Environments}
\author{Konstantin Dobler$^{*,\ddagger,\S}$}
\author{Simon Lehnerer$^{*,\dagger}$}
\author{\\Federico Scozzafava$^{\dagger}$}
\author{Jonathan Janke$^{\dagger}$}
\author{Mohamed Ali$^{\dagger}$}

\affiliation{$^\dagger$Apple}
\affiliation{$^\S$Hasso Plattner Institute \& ELLIS Unit Potsdam}
\affiliation{\\$^*$Equal contribution}
\affiliation{$^\ddagger$Work done during an internship at Apple}

\usepackage{catchfile}
\abstract{We present the \textbf{Multilingual Reasoning Gym}, an extension of Reasoning Gym~\citep{stojanovski2025reasoninggym}, that procedurally generates verifiable reasoning problems across \numlangs{} languages.
We translate templates for 94 tasks with native-speaker validation in 10 languages and targeted code or template adaptations to ensure linguistic naturalness. 
The {Multilingual Reasoning Gym} preserves the core benefits of the procedural generation approach used in the original Reasoning Gym, such as virtually unlimited problem instance generation and adjustable difficulty, and remains directly usable for Reinforcement Learning from Verifiable Rewards and evaluation settings. 
Problems in the Multilingual Reasoning Gym are parallel across languages, enabling crosslingually parallel data generation at massive scale due to the procedural nature of the environments.
 We release our implementation to support research into multilingual reasoning models.}

\newcommand{\promptgithub}{We provide the prompt templates on our GitHub.}

{
\metadata[Code \& Data]{
\url{https://github.com/apple/ml-multilingual-reasoning-gym}}
\metadata[Correspondence]{\sffamily konstantin.dobler@hpi.de}
\date{\sffamily March 11, 2026}
}
\begin{document}

\maketitle

\section{Introduction}

State-of-the-art large language models (LLMs) show that strong reasoning skills across math, logic, and algorithmic tasks are often trained using  Reinforcement Learning with Verifiable Rewards \citep[RLVR;][]{lambert2024tulu3}. The recently introduced Reasoning Gym \citep{stojanovski2025reasoninggym} provides a powerful framework for procedurally generating diverse reasoning problems at scale, offering unlimited training and evaluation data -- especially well-suited for RLVR. However, the original Reasoning Gym contains only English tasks and can thus not be leveraged for research and applications in other {languages}.

The ability to procedurally generate reasoning tasks offers unique advantages over static benchmarks and training datasets especially in the multilingual setting: (1) parametrizable templates allow for the generation of unlimited unique problem instances, whereas static multilingual datasets are limited in size, (2) controlled difficulty adjustment enables curriculum learning with tasks matching the current model's capabilities which can differ significantly across languages, and (3) instance generation across languages using the same seed allows for direct cross-lingual data at massive scale. However, extending procedural generation to multiple languages presents unique challenges beyond simple translation. Templates must be carefully translated to ensure that generated instances sound natural, notation and formatting conventions must be adapted, and the generation logic itself may need modification to accommodate linguistic differences.

\begin{table*}[t]
\centering
\small
\setlength{\tabcolsep}{6pt}
\begin{tabular}{p{0.34\textwidth} p{0.32\textwidth} p{0.26\textwidth}}
\toprule
\rowcolor{gray!10}
\textbf{English (Original)} & \textbf{Translated (Language)} & \textbf{Details} \\
\midrule
\textbf{Example 1 -- Number Sequence}\par
\textbf{Q:} {8, 14, 20, 26, 32, 38, 44, ?}\par
\textbf{A:} {50}
&
\textbf{Japanese}\par
\textbf{Q:} {8、14、20、26 、32、38、44、？}\par
\textbf{A:} {50}
&
Full-width punctuation translated (comma \texttt{, -> 、} and question mark \texttt{? -> ？}). \\
\midrule
\textbf{Example 2 -- Greatest Common Divisor}\par
\textbf{Q:} {Find the Greatest Common Divisor (GCD) of these numbers: 688, 716. Give only the GCD as your final answer.}\par
\textbf{A:} {4}
&
\textbf{German}\par
\textbf{Q:} {Bestimme den größten gemeinsamen Teiler (ggT) von diesen Zahlen: 688, 716. Gib nur den ggT als Antwort an.}\par
\textbf{A:} {4}
&Translated mathematical concepts using appropriate terminology commonly used in the target language (Greatest Common Divisor \texttt{GCD} $\rightarrow$ gr\"o\ss ter gemeinsamer Teiler \texttt{ggT}). \\
\midrule
\textbf{Example 3 -- String Isomorphism}\par
\textbf{Q:} {Return True if the following two strings are isomorphic, or False otherwise: zh lr}\par
\textbf{A:} {True}
&
\textbf{Italian}\par
\textbf{Q:} {Restituisci Vero se le seguenti due stringhe sono isomorfe, altrimenti Falso: zh lr}\par
\textbf{A:} {Vero}
&
Answer keywords translated (\texttt{True/False} $\rightarrow$ \texttt{Vero/Falso}). \\
\midrule
\textbf{Example 4 -- Bit Count}\par
\textbf{Q:} {How many 1 bits are there in the binary representation of the number 82789451?
}\par
\textbf{A:} {14}
&
\textbf{Japanese}\par
\textbf{Q:} 2進数における数値82789451の1ビットの個数を求めよ。
\par
\textbf{A:} {14}
&
Uses commonly understood translation for technical terms, e.g., ``binary'' $\rightarrow$ ``2進''. \\
\bottomrule
\end{tabular}
\caption{Examples of problems generated from translated templates and the corresponding English originals. Mathematical structure and verifiers remain identical; surface form (punctuation, terminology, answer tokens) is naturally translated.}
\label{tab:parallel_examples}
\vspace{-2mm}
\end{table*}

In this work, we present the Multilingual Reasoning Gym, a comprehensive extension of the Reasoning Gym to support multilingual procedural generation. We extend the Reasoning Gym's procedural generation framework with carefully translated task templates to cover 14 diverse languages in total. 
Our Multilingual Reasoning Gym supports English, Chinese, German, Spanish, French, Japanese, Brazilian Portuguese, Russian, Korean, Italian, Thai, Bengali, Telugu, and Swahili.
We use a hybrid human-LLM based process with native speaker validation ensuring natural and correct generated problem instances. We present selected samples in \autoref{tab:parallel_examples}. If necessary, we adapt task implementations to facilitate higher translation quality, especially for morphologically rich and non-Latin languages. 
Notably, our work preserves the core benefits of the Reasoning Gym -- unlimited reasoning problem generation with adjustable difficulty -- while extending this capability to multiple languages, enabling RLVR training and evaluation of multilingual reasoning models at scale.

\section{Related Work}

\paragraph{RLVR Environments.}
Reinforcement Learning with Verifiable Rewards (RLVR) has emerged as a practical recipe for training generative reasoning models by granting reward only when an automatic verifier validates the output. \citet{lambert2024tulu3} constructed an RLVR corpus from automatically gradable tasks -- e.g., GSM8K~\citep{cobbe2021trainingverifierssolvemath}, MATH~\citep{hendrycks2021measuringmathematicalproblemsolving} and IFEval~\citep{zhou2023instructionfollowingevaluationlargelanguage}. \citet{deepseekai2025deepseekr1incentivizingreasoningcapability} similarly scaled RL on verifiable math and science problems.
Focusing on symbolic reasoning, \citet{xie2025logicrlunleashingllmreasoning} procedurally generated formal logic puzzles and \citet{zhu2025autologiautomatedgenerationlogic} proposed synthesis of open-ended bilingual logic puzzles with verifiers. 
In parallel, efforts have been made to curate large-scale datasets: \citet{2025synthetic1} released \textsc{SYNTHETIC-1}, a 2M-questions suite with paired verifiers, and \citet{guha2025openthoughtsdatarecipesreasoning} aggregated open reasoning data (with verifiers) across domains.
Despite rapid progress, most RLVR environments, corpora, and public recipes remain English-centric. To our knowledge there is no large-scale, intrinsically multilingual RLVR environment with crosslingually parallel problems and verifiers; this gap motivates our work on the \emph{Multilingual Reasoning Gym}.

\paragraph{Multilingual Math and Reasoning Datasets.}
A multitude of multilingual reasoning benchmarks have been created primarily by translating fixed datasets. MGSM~\citep{shi2022languagemodelsmultilingualchainofthought} human-translates GSM8K items into ten languages. Recent math benchmarks expand language coverage and difficulty, such as MMATH \citep{luo2025mmathmultilingualbenchmarkmathematical}, PolyMath \citep{wang2025polymathevaluatingmathematicalreasoning} and MCLM \citep{son2025linguisticgeneralizabilitytesttimescaling}.
However, fixed datasets can cause overfitting or contamination~\citep{deng2024investigatingdatacontaminationmodern,xu2024benchmarkdatacontaminationlarge}, are by definition limited in size, and lack adaptive difficulty. In contrast, procedural generators offer unlimited problem instances with adaptive difficulty -- which is a desirable property for RLVR.

\section{The Multilingual Reasoning Gym}

\subsection{The English Reasoning Gym}
The Reasoning Gym \citep{stojanovski2025reasoninggym} is a library that procedurally generates problem statements, with accompanying verifiers, in over 100 different reasoning tasks spanning diverse domains including algebra, arithmetic, logic, and various games. 
Each task in the Reasoning Gym consists of a generator function that procedurally creates problem instances, a verifier that validates the solution correctness, and string templates that render parametrized problem instances into natural language.
This template-based approach offers a unique opportunity for multilingual dataset creation. While the original Reasoning Gym is solely in English, translating \textit{templates} instead of individual samples enables us to create virtually unlimited samples also in non-English languages, where data scarcity is usually much more pronounced. The translation effort scales only with the number of templates instead of the unlimited possible instantiations. Furthermore, due to the template-based approach, generated samples are parallel across all languages, enabling parallel data at a massive scale.
However, translating templates instead of individual samples also comes with particular challenges, especially when dealing with languages that have significantly different linguistic properties to English. Therefore, in this work, we provide carefully crafted translations as well as adaptations of existing tasks in the Reasoning Gym to create the \emph{Multilingual Reasoning Gym} with support for \numlangs{} languages. In the following, we describe the translation process.

\subsection{Translating templates to 13 non-English languages}

To ensure high-quality translated templates, we employ a hybrid approach of LLM-based machine translation, human native-speaker ratings as well as manual quality control. We now describe the steps we took to create high-quality translated template-based tasks for the Multilingual Reasoning Gym.

\paragraph{Cleaning \& filtering English-dependent tasks.}
As a first step, we clean and filter the tasks and templates in the original Reasoning Gym.
Some tasks in the English Reasoning Gym are difficult to translate (e.g., the task \texttt{games/word\_ladder} where the task is to build a chain of English words to transform one word into another by only changing a single letter). As the task is fundamentally bound to English and might not be possible in other languages (especially with non-Latin script), we decided to skip such tasks. Our implementation still supports the English version in this case. In total, we skip 10 such tasks out of the original 104 tasks.
Other tasks rely on data such as English example sentences, where words need to be reversed. In these cases, the task is not fundamentally bound to English but does use English data. We still translate these tasks but include that the data provided in the prompt will be English (e.g., ``Reverse words in the following \textit{English} sentence''). This case applies to a total of 6 tasks.
Finally, we make several small adaptations to the original templates and verification code for some tasks, alleviating verification errors or other flaws.
We provide details for each adapted task in \autoref{app:tasklist}.

\paragraph{Adapting Reasoning Gym to support multilingual sample generation.}
To facilitate procedural sample generation beyond English, we extract all hard-coded English strings into template JSON files, rewriting task code to support this if necessary.
To support languages other than English for a given task, it is thus only necessary to add a new JSON file that provides the translated templates, which makes it easy to extend the Multilingual Reasoning Gym to additional languages.
Furthermore, we adapt some tasks to make translations easier -- especially to morphologically rich languages. For example, some templates take advantage of properties of English such as being able to construct the plural of many nouns by simply appending an ``s''. Such simple constructions might not work in many other languages. When possible, we rewrite English prompts to side-step such issues,~e.g~by re-formatting constructions which require plurals ``\texttt{\{number\}} tiger\texttt{\{suffix\}}'' to ``tiger: \texttt{\{number\}}''. This affects a total of 14 tasks and we provide details on the changes made in \autoref{app:tasklist}.

\begin{table*}[h!]
\centering
\resizebox{\textwidth}{!}{%
\begin{tabular}{lccccccccccccccc}
\toprule
Model & en & zh & de & es & fr & it & pt & ru & ja & ko & th & bn & te & sw & Average \\
\midrule
gemma-3-1b-it & 4.7 & 4.4 & 3.6 & 4.5 & 3.6 & 4.2 & 4.5 & 5.6 & 2.5 & 2.3 & 4.4 & 3.6 & 2.8 & 1.6 & 3.7 \\
gemma-3-4b-it & 16.1 & 16.7 & 17.1 & 20.0 & 18.4 & 19.3 & 19.1 & 17.5 & 15.4 & 16.6 & 17.7 & 14.6 & 15.1 & 12.9 & 16.9 \\
gemma-3-12b-it & \textbf{23.2} & \textbf{19.7} & \textbf{19.1} & \textbf{20.9} & \textbf{20.6} & \textbf{21.4} & \textbf{19.8} & \textbf{20.1} & \textbf{19.2} & \textbf{19.5} & \textbf{19.7} & \textbf{17.5} & \textbf{17.7} & \textbf{17.7} & \textbf{19.7} \\
\midrule
Qwen3-1.7B & 28.4 & 28.2 & 30.3 & 30.8 & 30.3 & 29.2 & 29.4 & 28.8 & 30.4 & 25.9 & 27.0 & 24.6 & 23.0 & 9.9 & 26.9 \\
Qwen3-4B & 44.8 & 41.1 & 42.7 & 43.3 & 41.7 & 41.3 & 42.8 & 37.1 & 42.7 & 41.3 & 41.5 & 36.2 & 34.8 & 26.2 & 39.8 \\
Qwen3-8B & 45.7 & 43.4 & 46.3 & 45.6 & 45.4 & 45.1 & 44.7 & 41.5 & 46.0 & 43.9 & 43.3 & 39.2 & 37.9 & 32.3 & 42.9 \\
Qwen3-14B & \underline{\textbf{54.2}} & \underline{\textbf{48.3}} & \underline{\textbf{50.9}} & \underline{\textbf{50.2}} & \underline{\textbf{50.2}} & \underline{\textbf{50.3}} & \underline{\textbf{49.3}} & \textbf{45.2} & \underline{\textbf{50.4}} & \underline{\textbf{49.2}} & \underline{\textbf{49.2}} & \underline{\textbf{44.1}} & \underline{\textbf{42.3}} & \textbf{40.9} & \underline{\textbf{48.2}} \\
\midrule
Qwen3-1.7B-Base & 2.5 & 1.7 & 1.5 & 1.6 & 1.8 & 1.5 & 1.5 & 1.5 & 1.3 & 0.8 & 1.2 & 0.3 & 0.2 & 0.1 & 1.3 \\
Qwen3-4B-Base & 4.7 & 5.7 & 3.9 & 5.4 & 4.9 & 4.2 & 4.4 & 5.2 & 4.2 & 3.4 & 3.6 & 2.0 & 2.1 & 0.7 & 3.9 \\
Qwen3-8B-Base & 10.7 & 9.0 & 7.6 & 8.9 & 8.4 & 7.7 & 8.7 & 9.1 & 6.2 & 5.7 & 6.2 & 4.6 & 4.5 & 1.8 & 7.1 \\
Qwen3-14B-Base & \textbf{13.2} & \textbf{12.4} & \textbf{12.4} & \textbf{12.7} & \textbf{12.2} & \textbf{10.5} & \textbf{11.7} & \textbf{12.7} & \textbf{9.5} & \textbf{9.2} & \textbf{8.8} & \textbf{7.4} & \textbf{5.5} & \textbf{3.8} & \textbf{10.1} \\
\midrule
SmolLM3-3B & \textbf{21.0} & \textbf{13.3} & \textbf{17.3} & \textbf{15.2} & \textbf{17.8} & \textbf{17.2} & \textbf{17.7} & \textbf{17.1} & \textbf{14.2} & \textbf{9.7} & \textbf{12.7} & \textbf{8.2} & \textbf{3.9} & \textbf{4.2} & \textbf{13.5} \\
\midrule
gpt-oss-20b & \textbf{49.2} & \textbf{42.1} & \textbf{47.1} & \textbf{42.9} & \textbf{44.5} & \textbf{42.6} & \textbf{42.5} & \underline{\textbf{45.3}} & \textbf{43.5} & \textbf{36.7} & \textbf{42.8} & \textbf{41.9} & \textbf{40.6} & \underline{\textbf{41.2}} & \textbf{43.1} \\
\bottomrule
\end{tabular}
}
\caption{Average@8 scores of different models across languages on our Multilingual Reasoning Gym dataset at easy task difficulty (25th percentile of defined difficulty progressions). We use the recommended sampling parameters by the publisher for each model. Bold indicates best result within each model family. Underlined bold indicates overall best result.}
\label{tab:multilingual_results_easy}
\vspace{-2mm}
\end{table*}

\paragraph{Translation workflow.} After cleaning and filtering of the original 104 tasks, we are left with 94 tasks ready for translation (20 of which were adjusted or fixed as described above). 
In order to ensure high-quality translation, we make use of a hybrid strategy employing LLMs as well as human native speaker annotators, coupled with manual checking of templates and task logic by the authors. 

We first create an LLM-based translation using Claude Sonnet 4. Then, to improve the initial LLM-based translation, we employ an iterative LLM-based refinement process (also using Claude Sonnet 4). Here, translations are first graded using pre-defined rubrics and subsequently polished using a refinement prompt,\footnote{\promptgithub}
in case issues are detected. We observed that this approach is consistently able to catch obvious errors and allows us to provide better initial translations to our human annotators. Additionally, it allows us to integrate additional commonly found errors into the grading rubrics that are unearthed by our manual inspection or native speaker annotators.

The initial LLM-based translations are then graded by two human native speaker data annotators per sample in 10 languages (Chinese, German, French, Spanish, Italian, Russian, Thai, Japanese, Korean and Brazilian Portuguese). We do not run human native speaker validation for Bengali, Telugu and Swahili but provide translations for these languages as well to increase the coverage of underserved languages. As the underlying data generation process is template-based and sometimes requires reasoning through code for full understanding, we provide three procedurally generated samples per task to the human raters instead of showing the templates and accompanying code. We ask to identify logical issues and deviations from the original English samples as well as issues with word choice and general naturalness of the translations.

We then aggregate the task-specific as well as general feedback of the human annotators per language and across languages and re-run the LLM-based refinement process with improved and task-specific and language-specific guidelines. Finally, in all stages, we manually inspect the templates as well as the resulting procedurally generated samples to identify and fix issues.  We provide examples of final translations in \autoref{tab:parallel_examples}.

\section{Experimental Evaluation}

We evaluate various open-weight models on the Multilingual Reasoning Gym to demonstrate our dataset. For each task, we sample the same 50 problem instantiations across all languages and use the same data for all models. The original Reasoning Gym defines a per-task \textit{difficulty curriculum}. We use 25th percentile of each difficulty curriculum as our default setting (a lower-quartile difficulty level that avoids the easiest instances while remaining broadly solvable).
We report the average@8 evaluation results (average accuracy across 8 attempts per problem) in \autoref{tab:multilingual_results_easy}. 
Across all languages, performance increases with model capacity, indicating that the translated problems are solvable in principle given sufficiently strong models. At the same time, we observe substantial variation across languages, with English (en) consistently achieving the highest scores.

We therefore additionally evaluate a harder configuration using the 75th percentile of each difficulty curriculum (\autoref{app:difficult-evals}), which is substantially more challenging across all models and languages, reducing performance by up to 15 percentage points relative to our default setting. For completeness, we also report results using the original Reasoning Gym \textit{default} parameters (\autoref{app:default-diff-evals}), which include very easy instances and yield markedly higher scores. Between this easiest configuration and the harder 75th-percentile configuration, average performance can differ by up to 31 percentage points. These results demonstrate the ability of the Multilingual Reasoning Gym to cover a large range of difficulty levels.

\section{Conclusion}

We introduced the \textbf{Multilingual Reasoning Gym}, extending Reasoning Gym with carefully translated templates, supporting 94 tasks in \numlangs{} languages. Our design preserves the strengths of procedural generation -- unlimited problem instances, controllable difficulty, and parallel instances across languages -- while ensuring correctness and validity through extensive native-speaker validation. This makes the library immediately usable for multilingual RLVR training and cross-lingual model evaluation. We release our code and data to benefit the research community and facilitate future research into multilingual RLVR and reasoning models.

\section*{Limitations}
The Multilingual Reasoning Gym enables procedural reasoning problem generation in 14 languages. Three languages (Bengali, Telugu, and Swahili) were not validated by native speakers, so fluency and translation correctness may be less consistent for those settings. Not all tasks translate cleanly: some are omitted entirely (e.g., word games that rely on English-specific properties), and some translated tasks retain English source material (e.g., English words or sentences as inputs). As a result, these tasks are not fully multilingual in practice, and users may wish to filter them depending on their application. More broadly, template-based procedural generation does not fully represent real-world linguistic variation. Finally, our evaluation covers only a subset of open-weight models and a limited set of difficulty configurations.
\section*{Acknowledgments}
We thank the Reasoning Gym authors for creating the original framework that made this work possible. 
We further thank Adam Golinski and Arno Blaas for their insightful discussion and valuable feedback.
We also thank our team of professional translators and native speaker validators who ensured high-quality translations across all supported languages.
\bibliographystyle{plainnat}
\bibliography{custom}

\begin{thebibliography}{16}
\providecommand{\natexlab}[1]{#1}
\providecommand{\url}[1]{\texttt{#1}}
\expandafter\ifx\csname urlstyle\endcsname\relax
  \providecommand{\doi}[1]{doi: #1}\else
  \providecommand{\doi}{doi: \begingroup \urlstyle{rm}\Url}\fi

\bibitem[Cobbe et~al.(2021)Cobbe, Kosaraju, Bavarian, Chen, Jun, Kaiser, Plappert, Tworek, Hilton, Nakano, Hesse, and Schulman]{cobbe2021trainingverifierssolvemath}
Karl Cobbe, Vineet Kosaraju, Mohammad Bavarian, Mark Chen, Heewoo Jun, Lukasz Kaiser, Matthias Plappert, Jerry Tworek, Jacob Hilton, Reiichiro Nakano, Christopher Hesse, and John Schulman.
\newblock Training verifiers to solve math word problems, 2021.
\newblock URL \url{https://arxiv.org/abs/2110.14168}.

\bibitem[DeepSeek-AI et~al.(2025)DeepSeek-AI, Guo, Yang, Zhang, Song, Zhang, Xu, Zhu, Ma, Wang, Bi, Zhang, Yu, Wu, Wu, Gou, Shao, Li, Gao, Liu, Xue, Wang, Wu, Feng, Lu, Zhao, Deng, Zhang, Ruan, Dai, Chen, Ji, Li, Lin, Dai, Luo, Hao, Chen, Li, Zhang, Bao, Xu, Wang, Ding, Xin, Gao, Qu, Li, Guo, Li, Wang, Chen, Yuan, Qiu, Li, Cai, Ni, Liang, Chen, Dong, Hu, Gao, Guan, Huang, Yu, Wang, Zhang, Zhao, Wang, Zhang, Xu, Xia, Zhang, Zhang, Tang, Li, Wang, Li, Tian, Huang, Zhang, Wang, Chen, Du, Ge, Zhang, Pan, Wang, Chen, Jin, Chen, Lu, Zhou, Chen, Ye, Wang, Yu, Zhou, Pan, Li, Zhou, Wu, Ye, Yun, Pei, Sun, Wang, Zeng, Zhao, Liu, Liang, Gao, Yu, Zhang, Xiao, An, Liu, Wang, Chen, Nie, Cheng, Liu, Xie, Liu, Yang, Li, Su, Lin, Li, Jin, Shen, Chen, Sun, Wang, Song, Zhou, Wang, Shan, Li, Wang, Wei, Zhang, Xu, Li, Zhao, Sun, Wang, Yu, Zhang, Shi, Xiong, He, Piao, Wang, Tan, Ma, Liu, Guo, Ou, Wang, Gong, Zou, He, Xiong, Luo, You, Liu, Zhou, Zhu, Xu, Huang, Li, Zheng, Zhu, Ma, Tang, Zha, Yan, Ren, Ren, Sha, Fu, Xu, Xie, Zhang,
  Hao, Ma, Yan, Wu, Gu, Zhu, Liu, Li, Xie, Song, Pan, Huang, Xu, Zhang, and Zhang]{deepseekai2025deepseekr1incentivizingreasoningcapability}
DeepSeek-AI, Daya Guo, Dejian Yang, Haowei Zhang, Junxiao Song, Ruoyu Zhang, Runxin Xu, Qihao Zhu, Shirong Ma, Peiyi Wang, Xiao Bi, Xiaokang Zhang, Xingkai Yu, Yu~Wu, Z.~F. Wu, Zhibin Gou, Zhihong Shao, Zhuoshu Li, Ziyi Gao, Aixin Liu, Bing Xue, Bingxuan Wang, Bochao Wu, Bei Feng, Chengda Lu, Chenggang Zhao, Chengqi Deng, Chenyu Zhang, Chong Ruan, Damai Dai, Deli Chen, Dongjie Ji, Erhang Li, Fangyun Lin, Fucong Dai, Fuli Luo, Guangbo Hao, Guanting Chen, Guowei Li, H.~Zhang, Han Bao, Hanwei Xu, Haocheng Wang, Honghui Ding, Huajian Xin, Huazuo Gao, Hui Qu, Hui Li, Jianzhong Guo, Jiashi Li, Jiawei Wang, Jingchang Chen, Jingyang Yuan, Junjie Qiu, Junlong Li, J.~L. Cai, Jiaqi Ni, Jian Liang, Jin Chen, Kai Dong, Kai Hu, Kaige Gao, Kang Guan, Kexin Huang, Kuai Yu, Lean Wang, Lecong Zhang, Liang Zhao, Litong Wang, Liyue Zhang, Lei Xu, Leyi Xia, Mingchuan Zhang, Minghua Zhang, Minghui Tang, Meng Li, Miaojun Wang, Mingming Li, Ning Tian, Panpan Huang, Peng Zhang, Qiancheng Wang, Qinyu Chen, Qiushi Du, Ruiqi Ge, Ruisong
  Zhang, Ruizhe Pan, Runji Wang, R.~J. Chen, R.~L. Jin, Ruyi Chen, Shanghao Lu, Shangyan Zhou, Shanhuang Chen, Shengfeng Ye, Shiyu Wang, Shuiping Yu, Shunfeng Zhou, Shuting Pan, S.~S. Li, Shuang Zhou, Shaoqing Wu, Shengfeng Ye, Tao Yun, Tian Pei, Tianyu Sun, T.~Wang, Wangding Zeng, Wanjia Zhao, Wen Liu, Wenfeng Liang, Wenjun Gao, Wenqin Yu, Wentao Zhang, W.~L. Xiao, Wei An, Xiaodong Liu, Xiaohan Wang, Xiaokang Chen, Xiaotao Nie, Xin Cheng, Xin Liu, Xin Xie, Xingchao Liu, Xinyu Yang, Xinyuan Li, Xuecheng Su, Xuheng Lin, X.~Q. Li, Xiangyue Jin, Xiaojin Shen, Xiaosha Chen, Xiaowen Sun, Xiaoxiang Wang, Xinnan Song, Xinyi Zhou, Xianzu Wang, Xinxia Shan, Y.~K. Li, Y.~Q. Wang, Y.~X. Wei, Yang Zhang, Yanhong Xu, Yao Li, Yao Zhao, Yaofeng Sun, Yaohui Wang, Yi~Yu, Yichao Zhang, Yifan Shi, Yiliang Xiong, Ying He, Yishi Piao, Yisong Wang, Yixuan Tan, Yiyang Ma, Yiyuan Liu, Yongqiang Guo, Yuan Ou, Yuduan Wang, Yue Gong, Yuheng Zou, Yujia He, Yunfan Xiong, Yuxiang Luo, Yuxiang You, Yuxuan Liu, Yuyang Zhou, Y.~X. Zhu,
  Yanhong Xu, Yanping Huang, Yaohui Li, Yi~Zheng, Yuchen Zhu, Yunxian Ma, Ying Tang, Yukun Zha, Yuting Yan, Z.~Z. Ren, Zehui Ren, Zhangli Sha, Zhe Fu, Zhean Xu, Zhenda Xie, Zhengyan Zhang, Zhewen Hao, Zhicheng Ma, Zhigang Yan, Zhiyu Wu, Zihui Gu, Zijia Zhu, Zijun Liu, Zilin Li, Ziwei Xie, Ziyang Song, Zizheng Pan, Zhen Huang, Zhipeng Xu, Zhongyu Zhang, and Zhen Zhang.
\newblock Deepseek-r1: Incentivizing reasoning capability in llms via reinforcement learning, 2025.
\newblock URL \url{https://arxiv.org/abs/2501.12948}.

\bibitem[Deng et~al.(2024)Deng, Zhao, Tang, Gerstein, and Cohan]{deng2024investigatingdatacontaminationmodern}
Chunyuan Deng, Yilun Zhao, Xiangru Tang, Mark Gerstein, and Arman Cohan.
\newblock Investigating data contamination in modern benchmarks for large language models, 2024.
\newblock URL \url{https://arxiv.org/abs/2311.09783}.

\bibitem[Guha et~al.(2025)Guha, Marten, Keh, Raoof, Smyrnis, Bansal, Nezhurina, Mercat, Vu, Sprague, Suvarna, Feuer, Chen, Khan, Frankel, Grover, Choi, Muennighoff, Su, Zhao, Yang, Pimpalgaonkar, Sharma, Ji, Deng, Pratt, Ramanujan, Saad-Falcon, Li, Dave, Albalak, Arora, Wulfe, Hegde, Durrett, Oh, Bansal, Gabriel, Grover, Chang, Shankar, Gokaslan, Merrill, Hashimoto, Choi, Jitsev, Heckel, Sathiamoorthy, Dimakis, and Schmidt]{guha2025openthoughtsdatarecipesreasoning}
Etash Guha, Ryan Marten, Sedrick Keh, Negin Raoof, Georgios Smyrnis, Hritik Bansal, Marianna Nezhurina, Jean Mercat, Trung Vu, Zayne Sprague, Ashima Suvarna, Benjamin Feuer, Liangyu Chen, Zaid Khan, Eric Frankel, Sachin Grover, Caroline Choi, Niklas Muennighoff, Shiye Su, Wanjia Zhao, John Yang, Shreyas Pimpalgaonkar, Kartik Sharma, Charlie Cheng-Jie Ji, Yichuan Deng, Sarah Pratt, Vivek Ramanujan, Jon Saad-Falcon, Jeffrey Li, Achal Dave, Alon Albalak, Kushal Arora, Blake Wulfe, Chinmay Hegde, Greg Durrett, Sewoong Oh, Mohit Bansal, Saadia Gabriel, Aditya Grover, Kai-Wei Chang, Vaishaal Shankar, Aaron Gokaslan, Mike~A. Merrill, Tatsunori Hashimoto, Yejin Choi, Jenia Jitsev, Reinhard Heckel, Maheswaran Sathiamoorthy, Alexandros~G. Dimakis, and Ludwig Schmidt.
\newblock Openthoughts: Data recipes for reasoning models, 2025.
\newblock URL \url{https://arxiv.org/abs/2506.04178}.

\bibitem[Hendrycks et~al.(2021)Hendrycks, Burns, Kadavath, Arora, Basart, Tang, Song, and Steinhardt]{hendrycks2021measuringmathematicalproblemsolving}
Dan Hendrycks, Collin Burns, Saurav Kadavath, Akul Arora, Steven Basart, Eric Tang, Dawn Song, and Jacob Steinhardt.
\newblock Measuring mathematical problem solving with the math dataset, 2021.
\newblock URL \url{https://arxiv.org/abs/2103.03874}.

\bibitem[Lambert et~al.(2024)Lambert, Morrison, Pyatkin, Huang, Ivison, Brahman, Miranda, Liu, Dziri, Lyu, Gu, Malik, Graf, Hwang, Yang, Bras, Tafjord, Wilhelm, Soldaini, Smith, Wang, Dasigi, and Hajishirzi]{lambert2024tulu3}
Nathan Lambert, Jacob Morrison, Valentina Pyatkin, Shengyi Huang, Hamish Ivison, Faeze Brahman, Lester James~V. Miranda, Alisa Liu, Nouha Dziri, Shane Lyu, Yuling Gu, Saumya Malik, Victoria Graf, Jena~D. Hwang, Jiangjiang Yang, Ronan~Le Bras, Oyvind Tafjord, Chris Wilhelm, Luca Soldaini, Noah~A. Smith, Yizhong Wang, Pradeep Dasigi, and Hannaneh Hajishirzi.
\newblock Tülu 3: Pushing frontiers in open language model post-training.
\newblock 2024.

\bibitem[Luo et~al.(2025)Luo, Zhao, Sha, Wang, and Wen]{luo2025mmathmultilingualbenchmarkmathematical}
Wenyang Luo, Wayne~Xin Zhao, Jing Sha, Shijin Wang, and Ji-Rong Wen.
\newblock Mmath: A multilingual benchmark for mathematical reasoning, 2025.
\newblock URL \url{https://arxiv.org/abs/2505.19126}.

\bibitem[Mattern et~al.(2025)Mattern, Jaghouar, Basra, Straube, Ferrante, Gabriel, Ong, Weisser, and Hagemann]{2025synthetic1}
Justus Mattern, Sami Jaghouar, Manveer Basra, Jannik Straube, Matthew~Di Ferrante, Felix Gabriel, Jack~Min Ong, Vincent Weisser, and Johannes Hagemann.
\newblock Synthetic-1: Two million collaboratively generated reasoning traces from deepseek-r1, 2025.
\newblock URL \url{https://www.primeintellect.ai/blog/synthetic-1-release}.

\bibitem[Shi et~al.(2022)Shi, Suzgun, Freitag, Wang, Srivats, Vosoughi, Chung, Tay, Ruder, Zhou, Das, and Wei]{shi2022languagemodelsmultilingualchainofthought}
Freda Shi, Mirac Suzgun, Markus Freitag, Xuezhi Wang, Suraj Srivats, Soroush Vosoughi, Hyung~Won Chung, Yi~Tay, Sebastian Ruder, Denny Zhou, Dipanjan Das, and Jason Wei.
\newblock Language models are multilingual chain-of-thought reasoners, 2022.
\newblock URL \url{https://arxiv.org/abs/2210.03057}.

\bibitem[Son et~al.(2025)Son, Hong, Ko, and Thorne]{son2025linguisticgeneralizabilitytesttimescaling}
Guijin Son, Jiwoo Hong, Hyunwoo Ko, and James Thorne.
\newblock Linguistic generalizability of test-time scaling in mathematical reasoning, 2025.
\newblock URL \url{https://arxiv.org/abs/2502.17407}.

\bibitem[Stojanovski et~al.(2025)Stojanovski, Stanley, Sharratt, Jones, Adefioye, Kaddour, and Köpf]{stojanovski2025reasoninggym}
Zafir Stojanovski, Oliver Stanley, Joe Sharratt, Richard Jones, Abdulhakeem Adefioye, Jean Kaddour, and Andreas Köpf.
\newblock Reasoning gym: Reasoning environments for reinforcement learning with verifiable rewards, 2025.
\newblock URL \url{https://arxiv.org/abs/2505.24760}.

\bibitem[Wang et~al.(2025)Wang, Zhang, Tang, Wei, Yang, Wang, Sun, Sun, Zhang, Wu, Cang, Zhang, Huang, Lin, Huang, and Zhou]{wang2025polymathevaluatingmathematicalreasoning}
Yiming Wang, Pei Zhang, Jialong Tang, Haoran Wei, Baosong Yang, Rui Wang, Chenshu Sun, Feitong Sun, Jiran Zhang, Junxuan Wu, Qiqian Cang, Yichang Zhang, Fei Huang, Junyang Lin, Fei Huang, and Jingren Zhou.
\newblock Polymath: Evaluating mathematical reasoning in multilingual contexts, 2025.
\newblock URL \url{https://arxiv.org/abs/2504.18428}.

\bibitem[Xie et~al.(2025)Xie, Gao, Ren, Luo, Hong, Dai, Zhou, Qiu, Wu, and Luo]{xie2025logicrlunleashingllmreasoning}
Tian Xie, Zitian Gao, Qingnan Ren, Haoming Luo, Yuqian Hong, Bryan Dai, Joey Zhou, Kai Qiu, Zhirong Wu, and Chong Luo.
\newblock Logic-rl: Unleashing llm reasoning with rule-based reinforcement learning, 2025.
\newblock URL \url{https://arxiv.org/abs/2502.14768}.

\bibitem[Xu et~al.(2024)Xu, Guan, Greene, and Kechadi]{xu2024benchmarkdatacontaminationlarge}
Cheng Xu, Shuhao Guan, Derek Greene, and M-Tahar Kechadi.
\newblock Benchmark data contamination of large language models: A survey, 2024.
\newblock URL \url{https://arxiv.org/abs/2406.04244}.

\bibitem[Zhou et~al.(2023)Zhou, Lu, Mishra, Brahma, Basu, Luan, Zhou, and Hou]{zhou2023instructionfollowingevaluationlargelanguage}
Jeffrey Zhou, Tianjian Lu, Swaroop Mishra, Siddhartha Brahma, Sujoy Basu, Yi~Luan, Denny Zhou, and Le~Hou.
\newblock Instruction-following evaluation for large language models, 2023.
\newblock URL \url{https://arxiv.org/abs/2311.07911}.

\bibitem[Zhu et~al.(2025)Zhu, Huang, Peng, Lu, Yu, Cheng, Qiu, Huang, and Lin]{zhu2025autologiautomatedgenerationlogic}
Qin Zhu, Fei Huang, Runyu Peng, Keming Lu, Bowen Yu, Qinyuan Cheng, Xipeng Qiu, Xuanjing Huang, and Junyang Lin.
\newblock Autologi: Automated generation of logic puzzles for evaluating reasoning abilities of large language models, 2025.
\newblock URL \url{https://arxiv.org/abs/2502.16906}.

\end{thebibliography}

\appendix
\clearpage
\section{Additional evaluation results}
\label{app:additional-evals}

\subsection{Evaluation at 75th percentile of difficulty progression}
\label{app:difficult-evals}

\begin{table*}[h!]
\centering
\resizebox{\textwidth}{!}{%
\begin{tabular}{lccccccccccccccc}
\toprule
Model & en & zh & de & es & fr & it & pt & ru & ja & ko & th & bn & te & sw & Average \\
\midrule
gemma-3-1b-it & 3.5 & 2.7 & 2.9 & 3.6 & 2.4 & 3.3 & 2.9 & 3.7 & 1.5 & 1.2 & 2.2 & 1.9 & 1.7 & 0.6 & 2.4 \\
gemma-3-4b-it & 10.5 & 10.6 & 11.2 & 12.5 & 12.3 & 12.4 & 12.4 & 11.0 & 8.7 & 10.5 & 11.0 & 8.9 & 10.0 & 6.5 & 10.6 \\
gemma-3-12b-it & \textbf{14.9} & \textbf{12.6} & \textbf{12.2} & \textbf{14.0} & \textbf{12.9} & \textbf{13.4} & \textbf{12.6} & \textbf{12.7} & \textbf{11.5} & \textbf{12.4} & \textbf{12.1} & \textbf{10.4} & \textbf{11.7} & \textbf{11.0} & \textbf{12.4} \\
\midrule
Qwen3-1.7B & 17.3 & 18.0 & 20.3 & 20.0 & 19.4 & 18.5 & 18.4 & 18.4 & 20.3 & 17.3 & 17.4 & 16.1 & 15.2 & 5.3 & 17.3 \\
Qwen3-4B & 31.7 & 29.2 & 30.4 & 30.7 & 29.6 & 29.7 & 30.2 & 26.6 & 29.7 & 28.5 & 29.7 & 25.6 & 25.6 & 18.5 & 28.3 \\
Qwen3-8B & 30.3 & 30.0 & 32.2 & 31.4 & 32.4 & 31.8 & 30.5 & 29.2 & 31.3 & 29.8 & 30.2 & 27.1 & 26.5 & 22.1 & 29.6 \\
Qwen3-14B & \underline{\textbf{36.2}} & \underline{\textbf{33.6}} & \underline{\textbf{35.5}} & \underline{\textbf{34.3}} & \underline{\textbf{34.8}} & \underline{\textbf{33.9}} & \underline{\textbf{33.3}} & \textbf{31.2} & \underline{\textbf{34.7}} & \underline{\textbf{34.0}} & \underline{\textbf{33.9}} & \underline{\textbf{31.0}} & \textbf{28.5} & \underline{\textbf{28.7}} & \underline{\textbf{33.1}} \\
\midrule
Qwen3-1.7B-Base & 1.7 & 1.3 & 1.2 & 1.1 & 1.3 & 1.0 & 1.0 & 1.4 & 1.1 & 0.4 & 0.9 & 0.3 & 0.2 & 0.0 & 0.9 \\
Qwen3-4B-Base & 3.1 & 3.4 & 2.5 & 3.6 & 3.4 & 2.6 & 3.1 & 3.4 & 2.8 & 2.4 & 2.6 & 1.1 & 1.4 & 0.4 & 2.6 \\
Qwen3-8B-Base & 7.0 & 5.6 & 4.9 & 5.6 & 5.2 & 4.6 & 5.2 & 5.2 & 3.8 & 3.5 & 3.7 & 3.0 & 2.5 & 1.2 & 4.4 \\
Qwen3-14B-Base & \textbf{9.3} & \textbf{7.7} & \textbf{7.8} & \textbf{7.8} & \textbf{7.1} & \textbf{6.7} & \textbf{7.5} & \textbf{8.2} & \textbf{6.0} & \textbf{5.9} & \textbf{5.7} & \textbf{4.8} & \textbf{3.6} & \textbf{2.7} & \textbf{6.5} \\
\midrule
SmolLM3-3B & \textbf{13.5} & \textbf{8.8} & \textbf{12.0} & \textbf{10.1} & \textbf{11.4} & \textbf{11.3} & \textbf{11.4} & \textbf{10.7} & \textbf{9.0} & \textbf{6.2} & \textbf{8.3} & \textbf{4.7} & \textbf{2.3} & \textbf{1.8} & \textbf{8.7} \\
\midrule
gpt-oss-20b & \textbf{35.5} & \textbf{31.2} & \textbf{33.2} & \textbf{31.2} & \textbf{32.3} & \textbf{31.0} & \textbf{30.7} & \underline{\textbf{32.3}} & \textbf{31.3} & \textbf{27.1} & \textbf{30.1} & \textbf{29.9} & \underline{\textbf{29.0}} & \textbf{28.4} & \textbf{30.9} \\
\bottomrule
\end{tabular}
}
\caption{Average@8 scores of different models across languages on our Multilingual Reasoning Gym dataset at hard task difficulty (75th percentile of difficulty progression). We use the recommended sampling parameters by the publisher for each model. Bold indicates best result within each model family. Underlined bold indicates overall best result.}
\label{tab:multilingual_results_hard}
\vspace{-2mm}
\end{table*}

\subsection{Default difficulty parameters}
\label{app:default-diff-evals}

We additionally report evaluation results at the ``default'' difficulty level using the parameters defined by the original Reasoning Gym authors. These settings partially contain some very easy problem instances and we therefore see higher scores across all models.

To complement the Average@8 results reported in \autoref{tab:multilingual_results_avg}, we also report the pass@8 results in \autoref{tab:multilingual_results_pass}. 
Trends are mostly similar (with higher scores in general across the board) except for gpt-oss-20b which has relatively higher pass@8 compared to average@8 scores when compared to the Qwen3 (non-Base) model series. We also report results on \emph{language consistency}, which evaluates whether the reasoning and answer language used by the model matches the query language. 

We further manually inspect reasoning traces and answers and find that models of the \texttt{Qwen3}-reasoning family sometimes suffer from ``multilingual confusion'': for prompts which are not in English, Chinese or Russian, reasoning is usually done in English as shown in \autoref{tab:multilingual_results_language_consistency}. However for tasks which require a translated answer (e.g. German ``grün'' instead of English ``green''), we find that models sometimes do indeed arrive at the correct \textit{English} answer but do not attempt to translate it back to German, giving only the English version as an answer. This is graded as incorrect by the verification mechanism.

\begin{table*}
\centering
\resizebox{0.95\textwidth}{!}{%
\begin{tabular}{lccccccccccccccc}
\toprule
Model & en & zh & de & es & fr & it & pt & ru & ja & ko & th & bn & te & sw & Average \\
\midrule
gemma-3-1b-it & 7.2 & 7.7 & 5.1 & 6.2 & 6.2 & 5.8 & 7.2 & 7.6 & 4.3 & 4.0 & 6.4 & 6.1 & 4.2 & 2.3 & 5.7 \\
gemma-3-4b-it & 24.4 & 25.6 & 26.5 & 28.1 & 28.5 & 27.9 & 28.1 & 26.8 & 24.2 & 24.7 & 25.9 & 22.0 & 21.6 & 20.1 & 25.3 \\
gemma-3-12b-it & \textbf{33.1} & \textbf{30.1} & \textbf{28.7} & \textbf{30.1} & \textbf{30.1} & \textbf{30.4} & \textbf{29.3} & \textbf{29.2} & \textbf{29.4} & \textbf{29.5} & \textbf{29.8} & \textbf{26.4} & \textbf{26.3} & \textbf{26.8} & \textbf{29.2} \\
\midrule
Qwen3-1.7B & 41.3 & 38.9 & 42.1 & 42.8 & 43.2 & 41.0 & 41.6 & 40.0 & 43.1 & 36.6 & 38.5 & 31.4 & 30.3 & 12.3 & 37.4 \\
Qwen3-4B & 62.0 & 56.4 & 59.6 & 58.5 & 59.0 & 57.0 & 57.6 & 52.6 & 59.4 & 54.6 & 59.2 & 50.2 & 45.9 & 34.4 & 54.7 \\
Qwen3-8B & 63.1 & 59.7 & 63.9 & 61.4 & 62.1 & 61.7 & 61.4 & 57.1 & 63.0 & 59.7 & 60.3 & 55.4 & 54.3 & 45.8 & 59.2 \\
Qwen3-14B & \underline{\textbf{68.1}} & \underline{\textbf{63.5}} & \underline{\textbf{67.3}} & \underline{\textbf{65.8}} & \underline{\textbf{66.3}} & \underline{\textbf{66.0}} & \underline{\textbf{64.3}} & \underline{\textbf{62.9}} & \underline{\textbf{64.8}} & \underline{\textbf{64.1}} & \underline{\textbf{67.3}} & \underline{\textbf{62.7}} & \underline{\textbf{59.0}} & \underline{\textbf{55.4}} & \underline{\textbf{64.1}} \\
\midrule
Qwen3-1.7B-Base & 3.5 & 2.7 & 2.4 & 2.2 & 2.9 & 2.1 & 2.4 & 2.4 & 1.9 & 1.3 & 1.6 & 0.5 & 0.4 & 0.1 & 1.9 \\
Qwen3-4B-Base & 6.8 & 8.9 & 6.5 & 8.1 & 7.5 & 6.6 & 7.1 & 7.6 & 7.0 & 5.1 & 5.1 & 3.0 & 3.3 & 1.1 & 6.0 \\
Qwen3-8B-Base & 16.3 & 14.7 & 12.6 & 14.2 & 13.4 & 12.0 & 13.2 & 14.1 & 10.1 & 8.9 & 9.8 & 7.2 & 6.2 & 2.8 & 11.1 \\
Qwen3-14B-Base & \textbf{21.1} & \textbf{19.7} & \textbf{18.5} & \textbf{19.5} & \textbf{18.4} & \textbf{16.4} & \textbf{19.1} & \textbf{19.5} & \textbf{15.8} & \textbf{14.8} & \textbf{14.6} & \textbf{12.2} & \textbf{8.6} & \textbf{6.1} & \textbf{16.0} \\
\midrule
SmolLM3-3B & \textbf{31.4} & \textbf{22.8} & \textbf{27.9} & \textbf{23.0} & \textbf{27.2} & \textbf{25.6} & \textbf{25.9} & \textbf{25.0} & \textbf{21.7} & \textbf{14.9} & \textbf{19.0} & \textbf{10.8} & \textbf{4.8} & \textbf{6.0} & \textbf{20.4} \\
\midrule
gpt-oss-20b & \textbf{61.2} & \textbf{53.0} & \textbf{59.6} & \textbf{54.4} & \textbf{55.5} & \textbf{53.2} & \textbf{52.7} & \textbf{56.3} & \textbf{54.4} & \textbf{46.8} & \textbf{55.9} & \textbf{52.0} & \textbf{51.0} & \textbf{53.4} & \textbf{54.2} \\
\bottomrule
\end{tabular}
}
\caption{Average@8 scores of different models across languages on our Multilingual Reasoning Gym dataset at medium task difficulty. We use the recommended sampling parameters by the publisher for each model. Bold indicates best result within each model family. Underlined bold indicates overall best result.}
\label{tab:multilingual_results_avg}
\vspace{-2mm}
\end{table*}

\begin{table*}[h]
\centering
\resizebox{0.95\textwidth}{!}{%
\begin{tabular}{lccccccccccccccc}
\toprule
Model & en & zh & de & es & fr & it & pt & ru & ja & ko & th & bn & te & sw & Average \\
\midrule
gemma-3-1b-it & 21.2 & 22.3 & 18.9 & 21.5 & 21.1 & 19.3 & 21.7 & 22.8 & 16.2 & 16.4 & 19.4 & 20.1 & 15.1 & 9.3 & 18.9 \\
gemma-3-4b-it & 41.7 & 41.4 & 44.2 & 45.8 & 45.2 & 45.6 & 44.9 & 41.8 & 41.9 & 41.6 & 44.8 & 38.9 & 41.5 & 39.4 & 42.8 \\
gemma-3-12b-it & \textbf{54.5} & \textbf{49.3} & \textbf{51.6} & \textbf{53.0} & \textbf{51.2} & \textbf{52.1} & \textbf{51.2} & \textbf{49.4} & \textbf{50.8} & \textbf{51.1} & \textbf{51.6} & \textbf{45.8} & \textbf{47.0} & \textbf{49.2} & \textbf{50.6} \\
\midrule
Qwen3-1.7B & 61.4 & 58.6 & 60.5 & 61.8 & 62.1 & 58.7 & 60.3 & 58.7 & 62.6 & 56.2 & 58.3 & 52.6 & 50.5 & 26.6 & 56.4 \\
Qwen3-4B & 75.1 & 72.1 & 74.7 & 74.4 & 74.3 & 73.5 & 73.4 & 70.5 & 74.4 & 71.3 & 74.2 & 68.8 & 63.7 & 52.1 & 70.9 \\
Qwen3-8B & 76.8 & 75.0 & 77.5 & 75.8 & 76.7 & 76.1 & 76.8 & 74.2 & 76.3 & 74.7 & 75.6 & 72.7 & 68.9 & 65.3 & 74.5 \\
Qwen3-14B & \textbf{80.1} & \textbf{76.2} & \textbf{80.0} & \textbf{79.4} & \textbf{79.9} & \textbf{79.1} & \textbf{78.6} & \textbf{76.5} & \textbf{79.1} & \textbf{78.3} & \textbf{79.8} & \textbf{78.4} & \textbf{74.6} & \textbf{72.6} & \textbf{78.0} \\
\midrule
Qwen3-1.7B-Base & 17.9 & 14.3 & 13.9 & 12.6 & 15.4 & 12.2 & 13.3 & 14.2 & 11.3 & 8.0 & 9.8 & 3.8 & 2.6 & 0.5 & 10.7 \\
Qwen3-4B-Base & 30.5 & 34.6 & 28.3 & 33.3 & 31.2 & 30.3 & 30.5 & 32.1 & 29.8 & 25.6 & 25.1 & 17.1 & 16.7 & 7.1 & 26.6 \\
Qwen3-8B-Base & 45.1 & 44.4 & 41.3 & 41.2 & 40.9 & 38.2 & 41.6 & 41.2 & 35.9 & 32.0 & 34.3 & 29.8 & 26.5 & 15.7 & 36.3 \\
Qwen3-14B-Base & \textbf{51.1} & \textbf{49.2} & \textbf{48.0} & \textbf{49.2} & \textbf{47.9} & \textbf{46.3} & \textbf{47.5} & \textbf{49.8} & \textbf{45.5} & \textbf{42.7} & \textbf{44.2} & \textbf{39.1} & \textbf{32.6} & \textbf{27.2} & \textbf{44.3} \\
\midrule
SmolLM3-3B & \textbf{57.1} & \textbf{47.7} & \textbf{56.9} & \textbf{52.4} & \textbf{54.5} & \textbf{54.7} & \textbf{53.3} & \textbf{53.3} & \textbf{49.4} & \textbf{44.3} & \textbf{46.7} & \textbf{30.1} & \textbf{17.2} & \textbf{20.9} & \textbf{45.6} \\
\midrule
gpt-oss-20b & \underline{\textbf{84.0}} & \underline{\textbf{81.3}} & \underline{\textbf{83.5}} & \underline{\textbf{83.7}} & \underline{\textbf{83.1}} & \underline{\textbf{82.9}} & \underline{\textbf{82.5}} & \underline{\textbf{82.5}} & \underline{\textbf{83.5}} & \underline{\textbf{80.3}} & \underline{\textbf{82.9}} & \underline{\textbf{80.6}} & \underline{\textbf{79.3}} & \underline{\textbf{81.8}} & \underline{\textbf{82.3}} \\
\bottomrule
\end{tabular}
}
\caption{Pass@8 scores of different models across languages on our Multilingual Reasoning Gym dataset at medium task difficulty. We use the recommended sampling parameters by the publisher for each model. Bold indicates best result within each model family. Underlined bold indicates overall best result.}
\label{tab:multilingual_results_pass}
\end{table*}

\begin{table*}[htbp]
\centering
\resizebox{0.95\textwidth}{!}{%
\begin{tabular}{lccccccccccccccc}
\toprule
Model & en & zh & de & es & fr & it & pt & ru & ja & ko & th & bn & te & sw & Average \\
\midrule
gemma-3-1b-it & 82.5 & 5.4 & 25.3 & 17.2 & 23.6 & 24.0 & 16.2 & 23.6 & 12.2 & 13.8 & 18.6 & 24.6 & 21.0 & 10.0 & 22.7 \\
gemma-3-4b-it & \textbf{98.2} & \textbf{66.2} & \underline{\textbf{93.8}} & \underline{\textbf{76.1}} & \textbf{68.7} & \underline{\textbf{92.7}} & \underline{\textbf{91.1}} & \textbf{93.9} & \underline{\textbf{88.8}} & \underline{\textbf{84.1}} & \underline{\textbf{96.4}} & \underline{\textbf{92.3}} & \underline{\textbf{95.6}} & \underline{\textbf{76.8}} & \underline{\textbf{86.8}} \\
gemma-3-12b-it\textsuperscript{\textdagger} & 0.0 & 8.0 & 30.5 & 50.3 & 28.9 & 27.1 & 0.0 & 26.4 & 18.5 & 12.4 & 27.1 & 25.6 & 29.6 & 22.5 & 21.9 \\
\midrule
Qwen3-1.7B & 99.9 & 93.0 & 0.0 & 0.0 & 0.0 & 0.0 & 0.0 & 53.8 & \textbf{0.6} & 0.0 & 0.0 & 0.1 & 0.2 & \textbf{2.9} & 17.9 \\
Qwen3-4B & 100.0 & \underline{\textbf{93.7}} & 0.0 & 0.0 & 0.0 & 0.0 & 0.0 & \underline{\textbf{99.4}} & 0.1 & \textbf{0.0} & \textbf{0.1} & \textbf{2.1} & \textbf{0.4} & 0.1 & \textbf{21.1} \\
Qwen3-8B & 100.0 & 93.5 & \textbf{0.3} & \textbf{0.2} & \textbf{0.0} & \textbf{0.0} & \textbf{0.0} & 78.8 & 0.0 & 0.0 & 0.0 & 1.6 & 0.1 & 1.1 & 19.7 \\
Qwen3-14B & \underline{\textbf{100.0}} & 92.8 & 0.0 & 0.0 & 0.0 & 0.0 & 0.0 & 98.1 & 0.0 & 0.0 & 0.0 & 0.0 & 0.0 & 0.0 & 20.8 \\
\midrule
Qwen3-1.7B-Base & 73.6 & 37.9 & 30.8 & 32.1 & 35.8 & 33.6 & 34.0 & 36.1 & 45.4 & 32.9 & 42.9 & 53.9 & \textbf{49.2} & 48.1 & 41.9 \\
Qwen3-4B-Base & 83.1 & 51.3 & 49.0 & 42.4 & 42.1 & 45.1 & 44.9 & 58.0 & 49.9 & 52.9 & 37.5 & 50.2 & 41.3 & 48.2 & 49.7 \\
Qwen3-8B-Base & \textbf{95.3} & 62.5 & 69.8 & 72.6 & 66.1 & 70.5 & 69.3 & 75.5 & 60.3 & \textbf{57.3} & 63.4 & 73.5 & 43.9 & \textbf{61.2} & 67.2 \\
Qwen3-14B-Base & 92.2 & \textbf{72.5} & \textbf{74.3} & \textbf{75.3} & \underline{\textbf{75.9}} & \textbf{76.4} & \textbf{75.2} & \textbf{84.7} & \textbf{63.8} & 57.2 & \textbf{72.6} & \textbf{77.5} & 42.7 & 48.5 & \textbf{70.6} \\
\midrule
SmolLM3-3B & \textbf{100.0} & \textbf{85.5} & \textbf{0.3} & \textbf{0.2} & \textbf{0.4} & \textbf{0.9} & \textbf{0.2} & \textbf{8.7} & \textbf{4.3} & \textbf{2.4} & \textbf{3.3} & \textbf{0.4} & \textbf{1.5} & \textbf{0.8} & \textbf{14.9} \\
\midrule
gpt-oss-20b & \textbf{99.6} & \textbf{0.0} & \textbf{3.3} & \textbf{11.1} & \textbf{8.6} & \textbf{6.5} & \textbf{7.0} & \textbf{8.4} & \textbf{0.9} & \textbf{0.1} & \textbf{6.6} & \textbf{4.7} & \textbf{3.9} & \textbf{3.8} & \textbf{11.7} \\
\bottomrule
\end{tabular}
}
\caption{Language consistency scores (average@8) of different models across languages on our Multilingual Reasoning Gym dataset at medium task difficulty. We use the recommended sampling parameters by the publisher for each model. Bold indicates best result within each model family. Underlined bold indicates overall best result. \textsuperscript{\textdagger}: language consistency results are skewed for gemma-3-12b-it because the model continued to generate unrelated content after providing answers to the posed question.}
\label{tab:multilingual_results_language_consistency}
\end{table*}

\section{Further Examples}
We provide examples of tasks with remaining ``English content'' in the translated version in \autoref{tab:english_content_examples}. Finally, we provide examples of tasks which we do not translate in \autoref{tab:untranslated_tasks_examples}.

\begin{table*}[hbtp]
\centering
\small
\setlength{\tabcolsep}{6pt}
\begin{tabular}{p{0.34\textwidth} p{0.26\textwidth} p{0.36\textwidth}}
\toprule
\rowcolor{gray!10}
\textbf{English (Original)} & \textbf{Translated (Language)} & \textbf{Explanation} \\
\midrule
\textbf{Example 1 -- Spell Backward}\par
\textbf{Q:} {Spell this word backward (example: sun -> nus): palpation
}\par
\textbf{A:} {noitaplap}
&
\textbf{Spanish}\par
\textbf{Q:} Escribe esta palabra al revés (ejemplo: sol -> los): palpation
\par
\textbf{A:} {noitaplap}
&
The words are sampled from an English corpus but the task itself does not rely on any knowledge of the English language but rather \textit{just requires mechanistically reversing a sequence of letters}. Future work could add sufficiently large corpora for each language or use other generation techniques to generate letter sequences.  \\
\midrule
\textbf{Example 2 -- Letter Counting}\par
\textbf{Q:} {How many times does the letter ``w'' appear in the text: ``it into a watering place"?}\par
\textbf{A:} {1}
&
\textbf{Japanese}\par
\textbf{Q:} {テキスト「it into a watering place」の中に文字「w」は何回現れるか？}
\par
\textbf{A:} {1}
& The sentences are sampled from an English corpus but the task itself does not rely on any knowledge of the English language but rather \textit{just requires mechanistically counting characters}. Future work could add sufficiently large corpora for each language or use other generation techniques to generate letter sequences.\\
\midrule
\textbf{Example 3 -- Group Anagrams}\par
\textbf{Q:} {\texttt{[...]} Group the following list of words into anagrams:
\texttt{["escrod", "decors", "scored", "semitaur", "muriates"]}}\par
\textbf{A:} \texttt{[["decors", "escrod", "scored"], ["muriates", "semitaur"]]}
&
\textbf{German}\par
\textbf{Q:} {\texttt{[...]} Gruppiere die folgende Wortliste in Anagramme:
\texttt{["escrod", "decors", "scored", "semitaur", "muriates"]}}\par
\textbf{A:} \texttt{[["decors", "escrod", "scored"], ["muriates", "semitaur"]]}
&
The anagrams are sampled from an English corpus but the task itself does not rely on any knowledge of the English language but rather \textit{just requires mechanistically matching counts of characters}. Future work could add sufficiently large corpora for each language, however this is more involved than simply collecting a list of words as building a data-structure of anagrams is required. Additionally, in some languages, the concept of an anagram might not be commonly used (e.g. some non-Latin scripts).\\
\bottomrule
\end{tabular}
\caption{Examples of translated tasks which still contain some English language content. We list these separately for easy filtering. Note that for many tasks, the English content is not actually required to be understood semantically but simply serves as an input for string-based algorithmic tasks.}
\label{tab:english_content_examples}
\end{table*}

\begin{table*}[hbtp]
\centering
\small
\setlength{\tabcolsep}{6pt}
\begin{tabular}{p{0.48\textwidth} p{0.52\textwidth}}
\toprule
\rowcolor{gray!10}
\textbf{English (Not translated)} & \textbf{Reason} \\
\midrule
\textbf{Example 1 -- Word ladder}\par
\textbf{Q:} {Question: Transform the word ladder 'HAND' to 'GLEE' by changing one letter at a time.
Provide your answer as a comma-separated sequence of uppercase letters without spaces.
Each step must be a valid English word.}\par
\textbf{A:} \texttt{HAND,RAND,REND,FEND,FEED,FLED,FLEE,GLEE}

&
This task requires significant knowledge of English, so we cannot use it while keeping the English content as is. However, translation to other languages -- especially with non-Latin script -- poses a significant challenge, might even be impossible, or significantly alter the difficulty of the task.   \\
\midrule
\textbf{Example 2 -- Time Intervals}\par
\textbf{Q:} {A database query started at 05/12/2013 and ended at 06/12/2013. How long did the query take? Answer in D days.}\par
\textbf{A:} \texttt{31 days}

& Some templates in the task require disambiguation between US and European date formatting conventions (without providing such necessary context). In the example, the answer would be ``1 day'' if interpreted e.g. with European conventions (05/12/2013 -> 5th December 2013, 06/12/2013 -> 6th December 2013) instead of the ground truth ``31 days'' if interpreted in an US context (05/12/2013 -> 12th May 2013, 06/12/2013 -> 12th June 2013). Future work could insert such context or otherwise work on localizing the various date formatting conventions.\\
\midrule
\textbf{Example 3 -- Figlet Fonts}\par
\textbf{Q:} 
What word does this say?
\begin{verbatim}
.d88b.  888b.  .d88b     db     .d88b. 
8P  Y8  8  .8  8P       dPYb    YPwww. 
8b  d8  8wwK'  8b      dPwwYb       d8 
`Y88P'  8  Yb  `Y88P  dP    Yb  `Y88P'
\end{verbatim}
&

(1) The words are sampled from an English corpus -- localization would require constructing large-scale corpora for each language. This could be tackled in future work via existing dictionaries. (2) However, for languages using non-ASCII characters, the rendering algorithm to the ``ASCII-Art'' figlet fonts depicted would have to be adapted, which is non-trivial.\\
\bottomrule
\end{tabular}
\caption{Examples of tasks which were not translated due to various issues. Our implementation still supports the original English version.}
\label{tab:untranslated_tasks_examples}
\end{table*}

\clearpage
\section{Complete Task List}
\label{app:tasklist}
\begin{longtable}{p{1.8cm}p{5.5cm}p{2.6cm}p{3.5cm}}
\caption{Complete list of Reasoning Gym tasks and their translation status}
\label{tab:task_translation_status}\\
\toprule
\textbf{Category} & \textbf{Task} & \textbf{Translation Status} & \textbf{Notes} \\
\midrule
\endfirsthead

\multicolumn{4}{c}{\textit{Table \ref{tab:task_translation_status} continued from previous page}} \\
\toprule
\textbf{Category} & \textbf{Task} & \textbf{Translation Status} & \textbf{Notes} \\
\midrule
\endhead

\midrule
\multicolumn{4}{r}{\textit{Continued on next page}} \\
\endfoot

\bottomrule
\multicolumn{4}{l}{\textbf{Summary:} 74 fully translated, 20 Fully translated with adaptations, 10 not translated (104 total tasks)} \\
\bottomrule
\endlastfoot

\multicolumn{4}{c}{\cellcolor{gray!10}\textit{\textbf{Algebra (6 tasks)}}} \\
\midrule
Algebra & \texttt{complex\_arithmetic} & Fully translated & -- \\
Algebra & \texttt{intermediate\_integration} & Fully translated & -- \\
Algebra & \texttt{polynomial\_equations} & Fully translated with adaptation & Fixed rounding description error \\
Algebra & \texttt{polynomial\_multiplication} & Fully translated & Fixed non-determinism \\
Algebra & \texttt{simple\_equations} & Fully translated with adaptation & Fixed whitespace issue \\
Algebra & \texttt{simple\_integration} & Fully translated & -- \\

\midrule
\multicolumn{4}{c}{\cellcolor{gray!10}\textit{\textbf{Arithmetic (18 tasks)}}} \\
\midrule
Arithmetic & \texttt{basic\_arithmetic} & Fully translated & -- \\
Arithmetic & \texttt{bitwise\_arithmetic} & Fully translated & -- \\
Arithmetic & \texttt{calendar\_arithmetic} & Not translated & Date format translation is tricky to get right for all locales \\
Arithmetic & \texttt{chain\_sum} & Fully translated & -- \\
Arithmetic & \texttt{count\_bits} & Fully translated & -- \\
Arithmetic & \texttt{decimal\_arithmetic} & Fully translated & -- \\
Arithmetic & \texttt{decimal\_chain\_sum} & Fully translated & -- \\
Arithmetic & \texttt{dice} & Fully translated & -- \\
Arithmetic & \texttt{fraction\_simplification} & Fully translated & -- \\
Arithmetic & \texttt{gcd} & Fully translated & -- \\
Arithmetic & \texttt{gsm\_symbolic} & Not translated & Too many subtasks \\
Arithmetic & \texttt{lcm} & Fully translated & -- \\
Arithmetic & \texttt{leg\_counting} & Fully translated with adaptation & Format: ``animal: count'' to avoid pluralization \\
Arithmetic & \texttt{number\_format} & Fully translated & -- \\
Arithmetic & \texttt{power\_function} & Fully translated & -- \\
Arithmetic & \texttt{prime\_factorization} & Fully translated & -- \\
Arithmetic & \texttt{products} & Fully translated & -- \\
Arithmetic & \texttt{time\_intervals} & Not translated & Date format localization is tricky  \\

\midrule
\multicolumn{4}{c}{\cellcolor{gray!10}\textit{\textbf{Algorithmic (34 tasks)}}} \\
\midrule
Algorithmic & \texttt{ab} & Fully translated with adaptation & Removed unnecessary newline \\
Algorithmic & \texttt{base\_conversion} & Fully translated & -- \\
Algorithmic & \texttt{binary\_alternation} & Fully translated & -- \\
Algorithmic & \texttt{binary\_matrix} & Fully translated & -- \\
Algorithmic & \texttt{caesar\_cipher} & Fully translated with adaptation & Marked as \textbf{English} cipher text \\
Algorithmic & \texttt{count\_primes} & Fully translated & -- \\
Algorithmic & \texttt{cryptarithm} & Fully translated & -- \\
Algorithmic & \texttt{game\_of\_life} & Fully translated & -- \\
Algorithmic & \texttt{game\_of\_life\_halting} & Fully translated with adaptation & Fixed reward function bug and truncation bug for large boards \\
Algorithmic & \texttt{graph\_color} & Fully translated & -- \\
Algorithmic & \texttt{group\_anagrams} & Fully translated & -- \\
Algorithmic & \texttt{isomorphic\_strings} & Fully translated & Fixed non-determinism \\
Algorithmic & \texttt{jugs} & Fully translated with adaptation & Fixed JSON encoding \\
Algorithmic & \texttt{letter\_counting} & Fully translated & -- \\
Algorithmic & \texttt{letter\_jumble} & Fully translated with adaptation & Marked as \textbf{English} words \\
Algorithmic & \texttt{manipulate\_matrix} & Fully translated & -- \\
Algorithmic & \texttt{number\_filtering} & Fully translated & -- \\
Algorithmic & \texttt{number\_sorting} & Fully translated & -- \\
Algorithmic & \texttt{palindrome\_generation} & Fully translated & -- \\
Algorithmic & \texttt{palindrome\_partitioning} & Fully translated & -- \\
Algorithmic & \texttt{pool\_matrix} & Fully translated & -- \\
Algorithmic & \texttt{ransom\_note} & Fully translated & Fixed non-determinism \\
Algorithmic & \texttt{rotate\_matrix} & Fully translated & -- \\
Algorithmic & \texttt{rotten\_oranges} & Fully translated & -- \\
Algorithmic & \texttt{sentence\_reordering} & Fully translated with adaptation & Marked as \textbf{English} sentence \\
Algorithmic & \texttt{spell\_backward} & Fully translated & -- \\
Algorithmic & \texttt{spiral\_matrix} & Fully translated with adaptation & Fixed direction description \\
Algorithmic & \texttt{string\_insertion} & Fully translated & -- \\
Algorithmic & \texttt{string\_manipulation} & Fully translated & -- \\
Algorithmic & \texttt{string\_splitting} & Fully translated & -- \\
Algorithmic & \texttt{string\_synthesis} & Fully translated & -- \\
Algorithmic & \texttt{word\_ladder} & Not translated & English-specific task \\
Algorithmic & \texttt{word\_sequence\_reversal} & Fully translated & -- \\
Algorithmic & \texttt{word\_sorting} & Fully translated with adaptation & Marked as \textbf{English} words \\

\midrule
\multicolumn{4}{c}{\cellcolor{gray!10}\textit{\textbf{ARC (3 tasks)}}} \\
\midrule
ARC & \texttt{arc\_1d} & Fully translated with adaptation & Visual alignment adjusted \\
ARC & \texttt{arc\_agi} & Fully translated & -- \\
ARC & \texttt{rearc} & Fully translated & -- \\

\midrule
\multicolumn{4}{c}{\cellcolor{gray!10}\textit{\textbf{Code (2 tasks)}}} \\
\midrule
Code & \texttt{bf} & Fully translated & -- \\
Code & \texttt{codeio} & Not translated & External data dependencies \\

\midrule
\multicolumn{4}{c}{\cellcolor{gray!10}\textit{\textbf{Cognition (7 tasks)}}} \\
\midrule
Cognition & \texttt{color\_cube\_rotation} & Fully translated with adaptation & Reordered to avoid color declension \\
Cognition & \texttt{figlet\_font} & Not translated & Non-Latin script incompatible \\
Cognition & \texttt{modulo\_grid} & Fully translated & -- \\
Cognition & \texttt{needle\_haystack} & Not translated & Difficult to translate \\
Cognition & \texttt{number\_sequence} & Fully translated & -- \\
Cognition & \texttt{rectangle\_count} & Fully translated & -- \\
Cognition & \texttt{rubiks\_cube} & Fully translated & -- \\

\midrule
\multicolumn{4}{c}{\cellcolor{gray!10}\textit{\textbf{Games (17 tasks)}}} \\
\midrule
Games & \texttt{boxnet} & Fully translated with adaptation & English coordinates, formatting fixed \\
Games & \texttt{countdown} & Fully translated & -- \\
Games & \texttt{emoji\_mystery} & Fully translated with adaptation & Marked as \textbf{English} sentence \\
Games & \texttt{futoshiki} & Fully translated & -- \\
Games & \texttt{kakurasu} & Fully translated & -- \\
Games & \texttt{knight\_swap} & Fully translated & Fixed non-determinism \\
Games & \texttt{mahjong\_puzzle} & Fully translated with adaptation & Removed articles for better translation \\
Games & \texttt{maze} & Fully translated & -- \\
Games & \texttt{mini\_sudoku} & Fully translated & -- \\
Games & \texttt{n\_queens} & Fully translated & -- \\
Games & \texttt{puzzle24} & Fully translated & -- \\
Games & \texttt{rush\_hour} & Fully translated & -- \\
Games & \texttt{sokoban} & Fully translated with adaptation & Changed to arrow key notation \\
Games & \texttt{sudoku} & Fully translated & -- \\
Games & \texttt{survo} & Fully translated & -- \\
Games & \texttt{tower\_of\_hanoi} & Fully translated & -- \\
Games & \texttt{tsumego} & Fully translated & -- \\

\midrule
\multicolumn{4}{c}{\cellcolor{gray!10}\textit{\textbf{Geometry (2 tasks)}}} \\
\midrule
Geometry & \texttt{advanced\_geometry} & Fully translated & -- \\
Geometry & \texttt{simple\_geometry} & Fully translated & -- \\

\midrule
\multicolumn{4}{c}{\cellcolor{gray!10}\textit{\textbf{Graphs (5 tasks)}}} \\
\midrule
Graphs & \texttt{course\_schedule} & Fully translated with adaptation & Clarified prerequisite format \\
Graphs & \texttt{family\_relationships} & Not translated & Gender-specific, cultural context \\
Graphs & \texttt{largest\_island} & Fully translated & -- \\
Graphs & \texttt{quantum\_lock} & Fully translated & -- \\
Graphs & \texttt{shortest\_path} & Fully translated with adaptation & Use arrow keys to signify directions, fixed prompt \\

\midrule
\multicolumn{4}{c}{\cellcolor{gray!10}\textit{\textbf{Logic (7 tasks)}}} \\
\midrule
Logic & \texttt{aiw} & Fully translated & -- \\
Logic & \texttt{circuit\_logic} & Fully translated & -- \\
Logic & \texttt{knights\_knaves} & Not translated & Too many strings to translate \\
Logic & \texttt{propositional\_logic} & Fully translated with adaptation & Fixed typo (removed dot) \\
Logic & \texttt{self\_reference} & Fully translated & -- \\
Logic & \texttt{syllogism} & Fully translated with adaptation & Fix ``undistributed middle fallacy'' not being recognized as invalid syllogism, Use mathematical predicates like $\forall x \in A: x \in B$  instead of ``All A are B'' to avoid issues with quantifier agreements and cases in morphologically more complex languages than English \\
Logic & \texttt{zebra\_puzzles} & Not translated & Possible in principle, but too many individual strings and tricky case agreements etc. for some languages \\

\midrule
\multicolumn{4}{c}{\cellcolor{gray!10}\textit{\textbf{Induction (2 tasks)}}} \\
\midrule
Induction & \texttt{acre} & Not translated & -- \\
Induction & \texttt{list\_functions} & Fully translated & Just a list of numbers like ``\texttt{1, 2, 4, 8, ?}'', we adapt commas and question mark, e.g. for Chinese \\

\end{longtable}

\FloatBarrier

\applefootnote{ \textcolor{textgray}{\sffamily Apple and the Apple logo are trademarks of Apple Inc., registered in the U.S. and other countries and regions.}}

\end{document}